\documentclass[letterpaper, 10 pt, conference]{ieeeconf}  % Comment this line out if you need a4paper

\IEEEoverridecommandlockouts                              % This command is only needed if 
\usepackage{hyperref}       % hyperlinks
\usepackage{graphicx}
\usepackage{multirow}
\usepackage{subfig}

\title{\LARGE \bf
Multi-Modal Hybrid Architecture for Pedestrian Action Prediction
}
\author{Amir Rasouli*, Tiffany Yau, Mohsen Rohani and Jun Luo
\thanks{Authors are with Noah's Ark Laboratory,  Huawei, Markham, Canada.}
\thanks{* Corresponding author \tt\small amir.rasouli@huawei.com}}% <-this % stops a space

\begin{document}

\maketitle

\begin{abstract}
Pedestrian behavior prediction is one of the major challenges for intelligent driving systems in urban environments. Pedestrians often exhibit a wide range of behaviors and adequate interpretations of those depend on various sources of information such as pedestrian appearance, states of other road users, the environment layout, etc. To address this problem, we propose a novel multi-modal prediction algorithm that incorporates different sources of information captured from the environment to predict future crossing actions of pedestrians. The proposed model benefits from a hybrid learning architecture consisting of feedforward and recurrent networks for analyzing visual features of the environment and dynamics of the scene. Using the existing 2D pedestrian behavior benchmarks and a newly annotated 3D driving dataset, we show that our proposed model achieves state-of-the-art performance in pedestrian crossing prediction.
\end{abstract}

\section{Introduction}
Road user behavior prediction is one of the fundamental challenges for autonomous driving systems in urban environments. Predicting pedestrian behavior, in particular, is of great importance as they are among the most vulnerable road users and often exhibit a wide range of behaviors \cite{Rasouli_2017_IV} that are impacted by numerous environmental and social factors \cite{Rasouli_2019_ITS}. 

\begin{figure}
\vspace{+0.3cm}
\includegraphics[width=1\columnwidth]{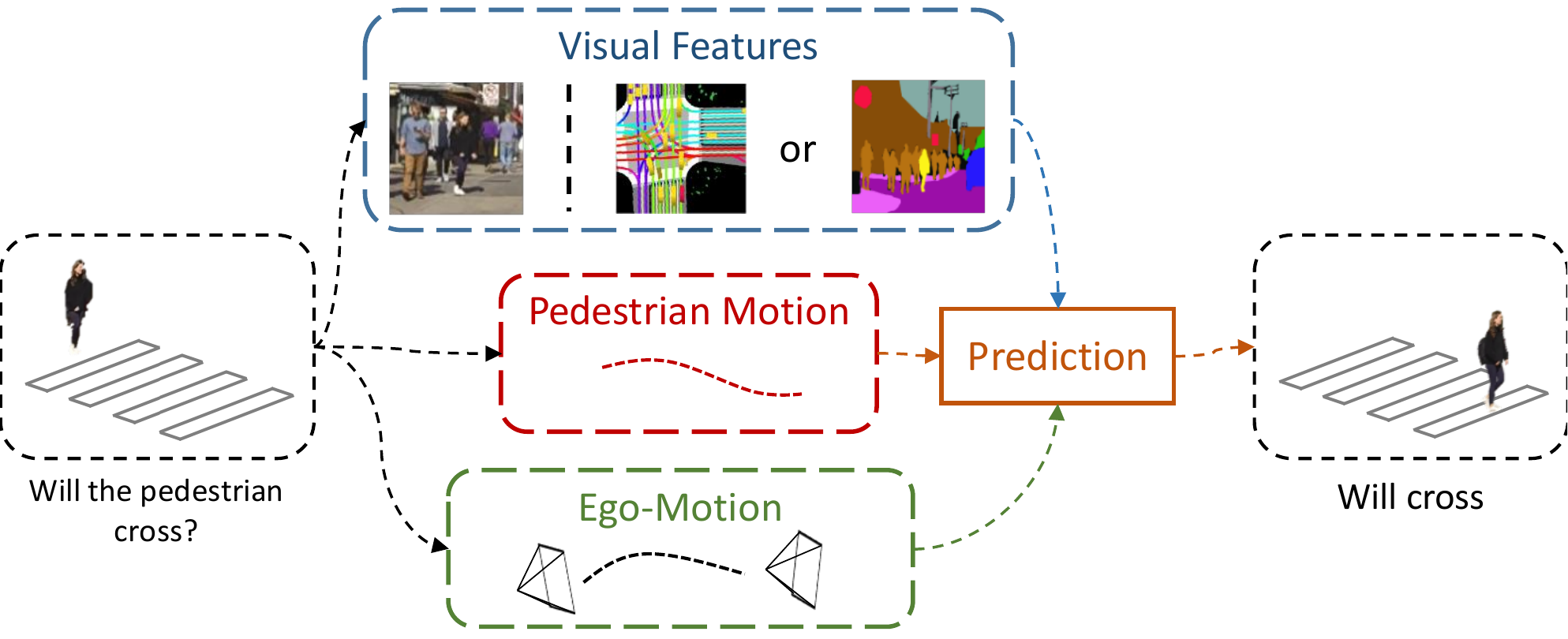}
\caption{Predicting pedestrian crossing in front of the ego-vehicle based on various contextual information: local scene around the pedestrian, semantic map of the environment and pedestrian and the ego-vehicle dynamics.}
\label{first_image}
\vspace{-0.5cm}
\end{figure}

Pedestrian prediction in the context of driving can be done implicitly via forecasting future trajectories or explicitly through predicting high-level actions of pedestrians, e.g. crossing the road (see Figure \ref{first_image}). In either case, behavior prediction requires a deep understanding of various contextual information in order to achieve a high precision \cite{Chaabane_2020_WACV, Malla_2020_CVPR, Rasouli_2019_BMVC,Rasouli_2019_ICCV,Liang_2019_CVPR,Aliakbarian_2018_ACCV}. In recent years, many deep learning architectures have been proposed that exploit various data modalities, such as visual features, pedestrian dynamics, pose, ego-motion, etc. in order to predict future actions. These algorithms rely on either feedforward methods for spatiotemporal reasoning over scene images or recurrent-based architectures to combine dynamics and visual features in a single framework (see \cite{Rasouli_2020_arxiv} for an extensive review).

In this work, we propose a novel hybrid architecture that benefits from both \textit{feedforward} and \textit{recurrent} architectures to generate a joint representation of both \textit{visual features} and \textit{scene dynamics}. More specifically, our model uses four modalities of data: semantic maps of the environment, local scenes of pedestrians and their surroundings, and  pedestrian and the ego-vehicle dynamics. The visual and dynamics features are encoded using convolutional layers and recurrent networks respectively. Their outputs are weighted using two attention modules and combined to create a joint representation that is used for action prediction. 

We evaluate the performance of the proposed algorithm on common 2D pedestrian behavior benchmark datasets as well as a new 3D dataset created by adding 3D bounding boxes and behavioral annotations to the existing autonomous driving dataset, nuScenes \cite{Caesar_2020_CVPR}. We show that the proposed method achieves state-of-the-art performance on multiple prediction metrics on these datasets.

\section{Related Works}
\subsection{Behavior prediction}
The topic of human behavior prediction is of great interest to robotics and computer vision communities. Behavior prediction has been used in many applications such as human-object \cite{Liu_2020_ECCV,Piergiovanni_2020_ECCV} and human-human interaction \cite{Joo_2019_CVPR,Yao_2018_CVPR}, risk assessment \cite{Strickland_2018_ICRA,Zeng_2017_CVPR}, anomaly detection \cite{Epstein_2020_CVPR}, surveillance \cite{Ma_2020_ECCV,Liang_2019_CVPR}, sports forecasting \cite{Qi_2020_CVPR,Felsen_2017_ICCV} and intelligent driving systems \cite{Fang_2020_CVPR,Rasouli_2019_BMVC}.

\subsection{Behavior Prediction in Driving}
In the context of intelligent driving, particularly in urban environments, predicting the behavior of road users is vital for safe motion planning. In this context, the dominant approach is to forecast future trajectories of other road users \cite{Fang_2020_CVPR,Liang_2020_CVPR,Phan-Minh_2020_CVPR,Chandra_2019_CVPR,Rhinehart_2018_ECCV}. Recent developments in the field, however, suggest that predicting higher-level actions of road users can be beneficial by providing an early risk assessment  \cite{Chaabane_2020_WACV,Rasouli_2019_BMVC,Gujjar_2019_ICRA} or by improving the accuracy of trajectory forecasting \cite{Malla_2020_CVPR,Rasouli_2019_ICCV,Liang_2019_CVPR,Casas_2018_CORL}. These actions, for example, can refer to various maneuvers performed by other vehicles, e.g. changing lane, making a turn \cite{Casas_2018_CORL}, or in the case of pedestrians, interaction with other objects, e.g. opening a car door \cite{Malla_2020_CVPR}, or, more importantly, crossing the road \cite{Rasouli_2019_ICCV,Rasouli_2019_BMVC,Gujjar_2019_ICRA}.

\subsubsection*{Pedestrian Crossing Prediction}
In recent years, the topic of pedestrian crossing prediction has gained momentum and many state-of-the-art algorithms have been proposed that take advantage of various data modalities and learning architectures \cite{Rasouli_2020_arxiv}.

Crossing prediction algorithms use different architectures. Those based on feedforward architectures \cite{Chaabane_2020_WACV,Saleh_2019_ICRA,Gujjar_2019_ICRA,Rasouli_2017_ICCVW} often rely on unimodal data. For instance, the method in \cite{Saleh_2019_ICRA} uses a series of 3D DenseNet blocks to reason over image sequences of pedestrians. The authors of \cite{Gujjar_2019_ICRA,Chaabane_2020_WACV} propose  generative encoder-decoder architectures using 3D convolutional layers. Here, the algorithms first generate future scenes and then based on those images predict crossing actions. The model in \cite{Rasouli_2017_ICCVW} relies on intermediate features, such as pedestrian head orientation, motion state, and the state of other traffic elements. Each of these components is detected using separate CNN-based object classifiers the outputs of which are combined for predicting crossing action.

Alternatively, some algorithms rely on recurrent architectures and take advantage of a combination of different modalities for prediction \cite{Liu_2020_RAL,Rasouli_2019_BMVC,Aliakbarian_2018_ACCV}. For instance, the authors of \cite{Aliakbarian_2018_ACCV} use a multi-stream LSTM architecture. They first encode visual features, optical flow images, and vehicle dynamics using individual LSTMs the outputs of which are concatenated to generate a shared representation. This representation is then fed into an embedding layer followed by another LSTM. The output of the second LSTM is combined with the shared representation for the final inference. The model proposed in \cite{Rasouli_2019_BMVC} uses a multi-layer recurrent architecture with five spatially stacked GRUs that encode pedestrian appearance, pedestrian surrounding context, pedestrians' poses and trajectories, and the ego-vehicle speed. The inputs are fed into the network at different levels according to their complexity, e.g. visual features at the bottom and speed at the top layers.

In this work, we use a hybrid approach that takes advantage of both feedforward networks and recurrent architectures for a joint visuospatial and dynamics reasoning over traffic scenes. 
 
\subsubsection*{Pedestrian Behavior Datasets} 
Although there are many publicly available datasets for pedestrian trajectory prediction  \cite{Liang_2020_CVPR_2,Sun_2020_CVPR_3,Liu_2018_CVPR,Robicquet_2016_ECCV,Lerner_2007_CGF}, the choices for pedestrian action prediction, particularly in the context of driving, are more limited.  There are a number of datasets that provide rich behavioral tags along with temporally coherent spatial annotations that can be used for pedestrian action prediction, including Joint Attention in Autonomous Driving (JAAD) \cite{Rasouli_2017_ICCVW}, VIrtual ENvironment for Action Analysis (VIENA$^2$) \cite{Aliakbarian_2018_ACCV}, Pedestrian Intention Estimation (PIE) \cite{Rasouli_2019_ICCV}, Trajectory Inference using Targeted Action priors Network (TITAN) \cite{Malla_2020_CVPR} and Stanford-TRI Intent Prediction (STIP) \cite{Liu_2020_RAL}. These datasets provide image sequences of traffic scenes recorded using on-board cameras along with 2D bounding boxes for pedestrians and other traffic objects as well as annotations for pedestrian behavior and ego-motion information. The drawback of these datasets is that they do not contain 3D information such as bird's eye view map of the environment, 3D bounding boxes, etc. all of which are important for real-world autonomous driving applications. In addition to 2D benchmark datasets, we use a new dataset of 3D bounding boxes and behavioral annotations that extends the nuScenes dataset.

\section{Method}

\subsection{Problem statement} In this work, pedestrian action prediction is formulated as an optimization problem  $p(A_i^{t+m}| LS, M,  PM, VM)$ where the goal is to estimate the probability of crossing action $A_i^{t+m} \in \{0,1\}$ for some pedestrian $i$, $1 < i < n$ in some time $t+m$ in the future. Here, the prediction is based on the observed local scene around the pedestrian  $LS = \{ls^{1}, ls^{2}, ..., ls^{t}\}$,  the semantic map of the environment $M = \{m^{1}, m^{2}, ..., m^{t}\}$, the pedestrian's motion $PM =\{pm^{1}, pm^{2}, ..., pm^{t}\}$ and the ego-vehicle motion $ VM = \{vm^{1}, vm^{2}, ..., vm^{t}\}$. 

\subsection{Architecture}
\begin{figure*}
\centering{
\includegraphics[width=0.8\textwidth]{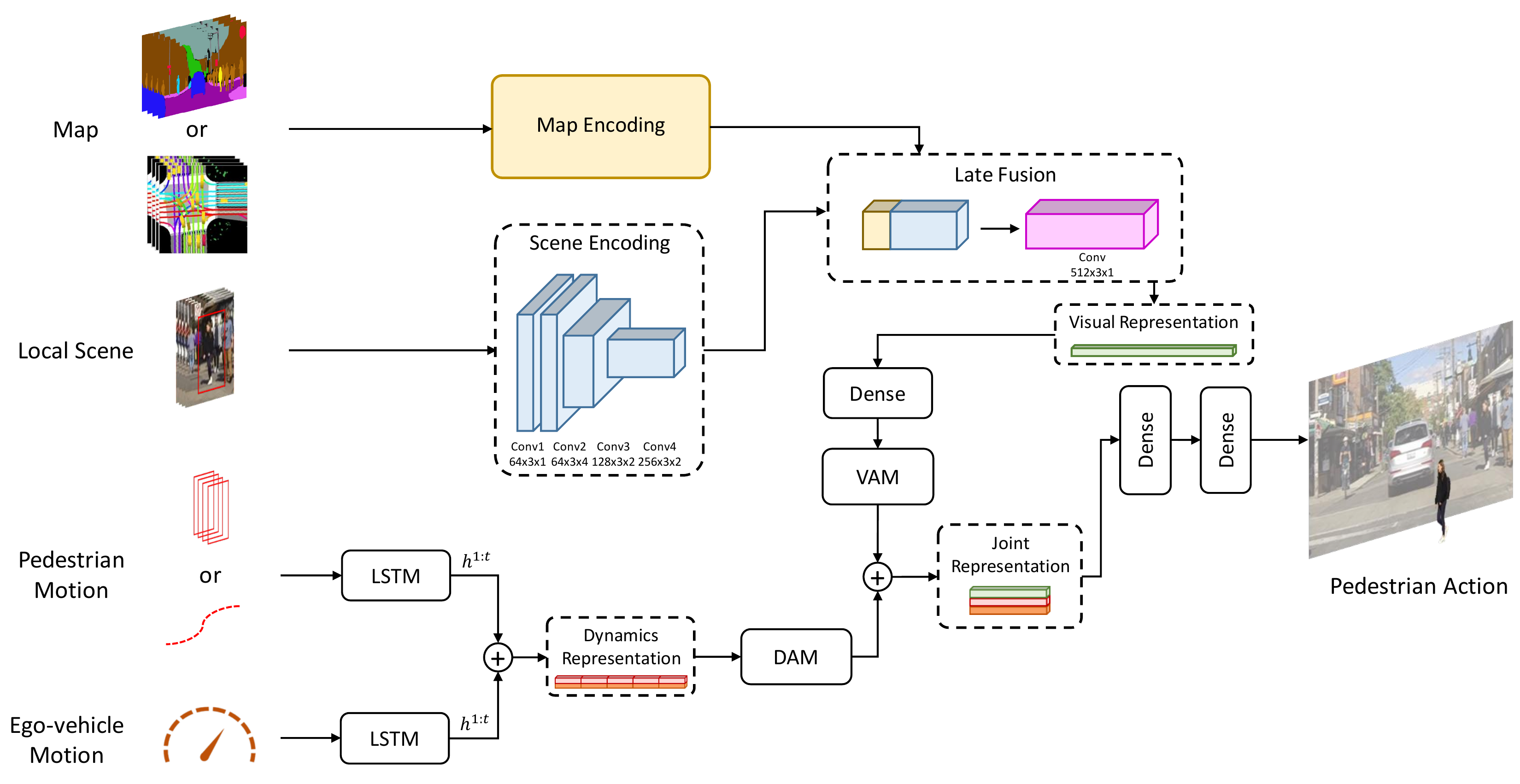}
\caption{The architecture of the proposed model. The model processes four different input modalities: semantic maps of the environment (2D or bird's eye view), local scene images of pedestrians and their motion (2D spatial or 3D global coordinates + velocity), and ego-vehicle motion. Semantic maps and images are encoded with two sets of Conv2D layers. The outputs of these layers are combined and fed into another Conv2D layer followed by a dense embedding layer to form visual representations which are weighted using Visual Attention Module (VAM). The motion of pedestrians and the ego-vehicle is encoded using two LSTMs, the outputs of which are concatenated and fed into Dynamic Attention Module (DAM). The final joint representation is formed by concatenating the outputs of the attention modules. The specifications of conv layers for scene encoding are given as \textit{[number of filters, kernel size, stride]}.}
\label{main_diagram}}
\vspace{-0.8cm}
\end{figure*}

As highlighted in a survey of the past studies \cite{Rasouli_2019_ITS}, there is a diverse set of social and environmental factors that impact the way pedestrians make crossing decision. To capture such contextual complexity, we propose a multi-modal method that relies on a hybrid architecture for encoding visual and dynamics information. Figure \ref{main_diagram} illustrates the proposed method which can be divided into three main parts:
\begin{itemize}
\item \textbf{Visual Encoding.} This part of the model relies on 2D convolutional layers to generate visual representations for \textit{semantic map} of the environment and \textit{local scene} context of pedestrians.
\item \textbf{Dynamics Encoding.} This recurrent part of the model generates a joint representation of scene dynamics, namely \textit{pedestrian} and \textit{ego-vehicle} motion. 
\item \textbf{Prediction.} Using a joint representation of the visual and dynamics features, the model estimates the probability of the crossing action in the future.
\end{itemize}
We discuss each of these components in more details in the subsections below.

\subsection{Visual Encoding}
\begin{figure}
\centering{
\subfloat[Sequential]{\includegraphics[width=0.30\columnwidth]{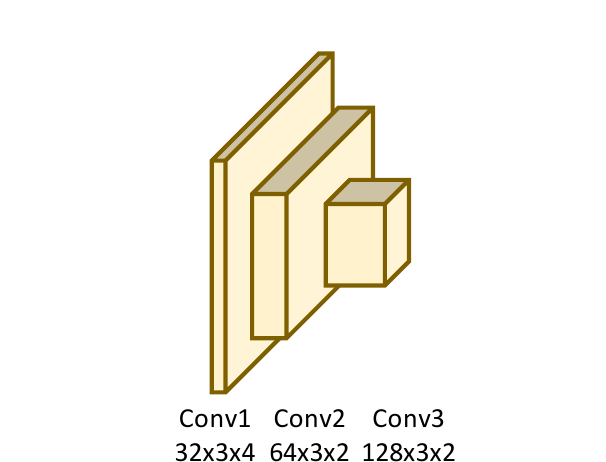}}
\hspace{0.1cm}
\subfloat[Atrous]{\includegraphics[width=0.30\columnwidth]{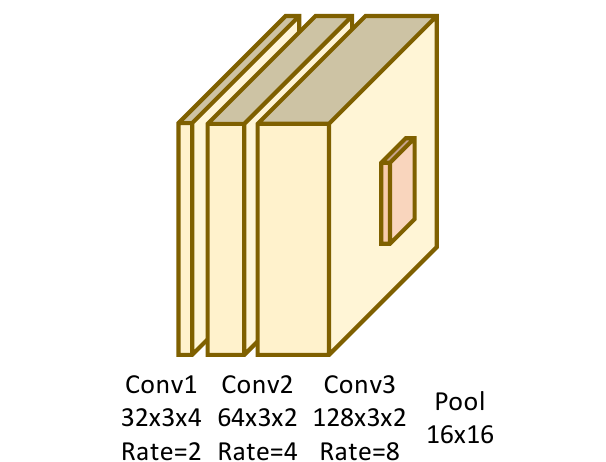}}
\hspace{0.1cm}
\subfloat[Multi-scale]{\includegraphics[width=0.30\columnwidth]{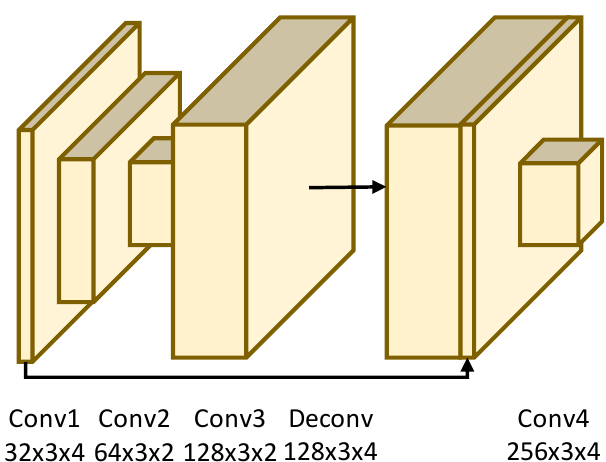}}
\caption{Different strategies for processing map images. The specifications of 2D conv layers are given as \textit{[number of filters, kernel size, stride]}. In b), rate refers to the rate of kernel dilation.}
\label{map_encoding}}
\vspace{-0.5cm}
\end{figure}

Visual information is crucial for robust prediction of pedestrian behavior as it reflects the state of the pedestrians, e.g. pose, head orientation, motion, gestures as well as their surrounding environment, e.g. the state of other road users, road structure, signals, etc. For this purpose, we rely on two different sources of visuospatial inputs: semantic maps and local scene images around pedestrians.

\textbf{Semantic map.} Depending on the type of data, we either use semantic maps generated from scene images or bird's eye view maps of the environment and traffic elements. The maps for each time-step are concatenated channel-wise.

When processing the maps, it is important to capture the interdependency between different traffic components such as pedestrians, vehicles, road structure, etc. To address this issue, we consider three strategies as shown in Figure \ref{map_encoding}. 

The first approach is a common sequential processing with successive convolutional layers. This method captures local information in the image but fails to represent broader context at higher resolutions. Furthermore, consecutive downsampling can result in the loss of some of the finer visual details. To address these drawbacks, we consider two alternatives: \textit{Atrous} (or dilated) convolution and \textit{multi-scale} skip connection. 

Atrous convolution is a special form of convolution in which the effective field of view of a filter is increased by inserting gaps, parameterized as the rate of dilation, between the elements of a given kernel as proposed in \cite{Chen_2017_arxiv}. This allows the model to capture a broader spatial context without the need for downsampling.

Multi-scale skip connections are often used to combine fine and coarse features generated at different convolutional layers. Following this approach, the output of the first conv layer and the upsampled (via deconvolution operation) output of the last layer are fed into another conv layer for joint processing. The specifications are shown in Figure \ref{map_encoding}.

\textbf{Local scene.} We use scene images to capture the changes in the appearance of pedestrians and their local surroundings. At a given time step, we extract a region around each pedestrian by scaling up the corresponding 2D spatial bounding box  and then adjust the dimensions of the scaled box so its width matches its height. Similar to maps, we concatenate the images channel-wise for a given observation sequence prior to processing by multiple conv layers.

In order to combine the visual features with different modalities, we employ a late fusion technique in which we spatially concatenate the outputs of map and local scene encodings and jointly process them with an additional convolutional layer. The output of the fusion conv layer is flattened and fed into a dense layer to reduce the dimension of the feature vector.

\textbf{Visual Attention Module (VAM).} Although visual features are important for reasoning about future pedestrian action, it is important to emphasize those features that are more relevant to the task. To achieve this goal, we weight visual representation features by using an attention mechanism. We calculate feature weight $\alpha_i$ for a given feature point $z_i$ as follows:

$$ \alpha_i = \frac{e^{f_i}}{\sum_j e^{f_j}},  f_i = \sum_k \omega_{i,k}z_k + b_i . $$

\noindent where $f_i$ is a single neuron in a fully-connected layer.
\subsection{Dynamics Encoding}
\label{sec:dynamics}

When anticipating pedestrian action, dynamics features, such as pedestrian and ego-motion, are of vital importance. For example, when a pedestrian is moving towards the road, that can be an indicator of crossing intention. Likewise, the ego-vehicle motion can directly impact whether a pedestrian makes a crossing decision. If, for instance, the ego-vehicle is driving at a high speed, the pedestrian will likely cross after the vehicle has passed. Given the importance of dynamics information, we use pedestrian and the ego-vehicle motion as described below.

\textbf{Pedestrian motion.} We use both pedestrian location and velocity for encoding pedestrian motion. In the absence of 3D data,  2D bounding boxes, i.e. the coordinates of the boxes $[x_1,y_1,x_2,y_2]$ corresponding to top-left and bottom-right corners of the boxes around pedestrians in image frames are used. For the 3D dataset, we use global coordinates of pedestrians in 3D space with respect to the map coordinates.

Velocities are obtained by calculating the displacement of the pedestrian coordinates (either in 2D or 3D) with respect to the first coordinate at the beginning of observation sequences.  The combination of pedestrian locations and velocities forms pedestrian motion information.

\textbf{Ego-vehicle motion.} We follow the same routine as for pedestrians and use the ego-vehicle global coordinates and velocities. For 2D datasets, since ego-location is not available, we only use the vehicle's velocity.

Pedestrian and ego-motion features are encoded using two recurrent networks. The hidden states  $h^{1:t}$  of these networks are concatenated temporally to form dynamics representations.   

\textbf{Dynamics Attention Module (DAM).} Similar to the visual representations, we employ an attention module to weight dynamics features. DAM is a 3D attention mechanism inspired by \cite{Luong_2015_arxiv} which receives as input a temporal data sequence and generates a unified weighted representation computed as follows: 

$$DAM^t = tanh(W_c[c^t\oplus h^t])$$
 $$ c^t = \sum_{i \in [1:t]}\sigma({h^t}' W_a h^i)h^i$$

\noindent  $\sigma$ is a softmax function and $h^i$ is a hidden state of  dynamics representation for $ i = 1,2,..., t$. Here, $\oplus$ denotes the concatenation operation.
 
\subsection{Prediction}
The joint representation used for final prediction is formed by concatenating the outputs of VAM and DAM. This representation is then fed into two consecutive dense layers to make crossing predictions. For learning, we use binary cross-entropy loss given by

$$L = - \sum_n y_n\log{\hat{y}_n} + (1-y_n)\log{(1-\hat{y}_n)}.$$

\section{Empirical Evaluation}

\subsection{Datasets}
For evaluations on 2D data, we choose two naturalistic pedestrian behavior benchmarks, PIE \cite{Rasouli_2019_ICCV}, and JAAD \cite{Rasouli_2017_ICCVW} which are publicly available. The alternatives, TITAN \cite{Malla_2020_CVPR} and STIP \cite{Liu_2020_RAL}  datasets are only available under restrictive terms of use, and VIENA$^2$ \cite{Aliakbarian_2018_ACCV} only contains simulated samples. 

\textbf{Pedestrian Intention Estimation (PIE).}  PIE contains 6 hours of driving footage recorded with an on-board camera. There are 1842 pedestrian samples with dense 2D bounding box annotations at 30Hz with behavioral tags along with motion sensor recordings both from the vehicle and the camera. For the ego-vehicle velocity, we use gyroscope readings from the camera. The data is split into training and testing sets as recommended in \cite{Rasouli_2019_ICCV}.

\textbf{Joint Attention in Autonomous Driving (JAAD).}  Similar to PIE, JAAD is a dataset consisting of on-board video recordings with 346 clips and 2580 pedestrians annotated with 2D bounding boxes at 30Hz. Compared to PIE, JAAD samples are less diversified, i.e. the majority of samples are pedestrians walking on sidewalks and sequences are generally shorter. JAAD also lacks ego-motion information, however, it provides driver's actions, such as \textit{speeding up}, \textit{stopping}, etc. which are used in place for ego-motion information. The data is split into train/test sets following \cite{Rasouli_2018_ECCVW}.

\textbf{Pedestrian Prediction on nuScenes (PePScenes).} Due to the lack of 3D driving data with pedestrian behavior annotations, we created our own dataset by adding annotations to the existing dataset nuScenes \cite{Caesar_2020_CVPR}. For behavioral annotations, we added sample-wise and per-frame crossing labels to 719 pedestrian samples out of which 570 are not crossing. For behavioral annotations, we selected the pedestrians that 1) appeared long enough in front of the vehicle and 2) appeared to have the intention of crossing, e.g. moving close to the curbside, looking at the traffic, etc. 

In addition, we extended 2D/3D bounding box annotations on nuScenes from 2Hz to 10Hz to make samples more suitable for pedestrian prediction. This is done by interpolating boxes between two consecutive frames using pedestrian global coordinates. The ratio of train/test data is 70/30.

\textbf{Data preparation.} We follow the same procedure as in \cite{Rasouli_2019_BMVC} and clip sequences up to the time of the crossing events, i.e. the moment the pedestrians start crossing the road. In cases where no crossing occurs, the last frame is selected instead. All models use $0.5$s observation length (or $5$ frames at $10$Hz for all datasets) and sample sequences from each track between $1$ to $2s$ to the event of crossing with $50\%$ overlap between each sample. Overall, we get the following number of train/test samples: $3980/3185$ for PIE, $3955/3110$ for JAAD and $2544/1146$ for PePScenes.

\subsection{Implementation}

For 2D datasets, we generate semantic maps using the algorithm in \cite{Chen_2017_arxiv} pretrained on Cityscapes \cite{Cordts_2016_CVPR} and use 14 main object classes appearing in traffic scenes. For the PePScenes dataset, rasterized maps encoded as 3-channel images similar to \cite{Cui_2019_ICRA} are used. The map is of size $30 \times 30$ meters centered around the ego-vehicle.

To extract local scene images, we use $1.5x$ scaled versions of pedestrian 2D bounding boxes in image frames. Unlike the 2D datasets, PePScenes contains recordings from three front looking cameras that cover a wide-angle view in front of the vehicle. To select a pedestrian of interest,  we choose a camera frame in which the pedestrian appears. If the pedestrian appears in two adjacent cameras, we select the camera frame in which a larger portion of the pedestrian is visible. 

The details of convolutional layers for each module can be seen in Figures \ref{main_diagram},\ref{map_encoding}. For all recurrent networks, we use LSTM cells with 256 hidden units. We set the dimension of the embedding dense layer for visual features to $512$ and the second to last layer to $256$.

\subsection{Training}
We trained the proposed model on all datasets end-to-end using RMSProp \cite{Tieleman_2012_tech} optimizer with batch size of $16$, learning rate of $0.00005$ for $50$ epochs on the 2D datasets and $40$ on PePScenes. We applied $L2$ regularization of $0.0001$ to LSTMs and the last dense layer. To compensate for data imbalance, we set class weights based on the ratio of positive and negative samples.

\subsection{Metrics}
To evaluate the proposed model, common binary classification metrics as in \cite{Rasouli_2019_BMVC} are used including $accuracy$, Area Under the Curve ($AUC$), $F1$, and $precision$. Using all these metrics, as opposed to only reporting on accuracy \cite{Liu_2020_RAL}, is important given the fact that the number of negative and positive samples are imbalanced in all the datasets mentioned above. This means that the models can favor one class over the others and still achieve very high accuracy. AUC, F1, and precision show the balanced performance of the methods by highlighting how on average each method identifies different classes and distinguishes between them.

\subsection{Models}
We compare the performance of the proposed model to a series of baseline and state-of-the-art crossing prediction models:\\
\textbf{Trajectory-Based Forecaster (TF)}. As a baseline, we make crossing prediction using only pedestrian trajectory information, i.e. pedestrian coordinates in 2D or 3D  (as discussed in Sec. \ref{sec:dynamics}), using an LSTM followed by a dense layer. \\
\textbf{Inception 3D (I3D)} \cite{Carreira_2017_CVPR}. Given the similarities between action prediction and recognition tasks, we use a state-of-the-art activity recognition model, I3D. This is a feedforward model based on a series of 3D convolutional layers. As input to this model, we use local scene images as described earlier.\\
\textbf{Stacked with Fusion GRU (SF-GRU)} \cite{Rasouli_2019_BMVC}. This is a state-of-the-art pedestrian crossing prediction algorithm based on a multi-level recurrent architecture in which different modalities of data are infused and encoded gradually. This model relies on five modalities of data, namely pedestrians' appearances and surrounding context, pedestrians' poses and coordinates, and the ego-vehicle speed.  We also report on a variation of SF-GRU in which we combine all data modalities and feed them to the bottom layer of a spatially \textbf{Stacked GRUs (S-GRU)} similar to \cite{Yue_2015_CVPR}. \\
An alternative approach to stacked processing is to encode different modalities of data individually and combine them prior to classification. We follow the approach in \cite{Aliakbarian_2018_ACCV} and process each input using a separate GRU without the second GRU for joint processing. We refer to this model as \textbf{Multi-Stream GRU (MS-GRU)}.

All the poses are generated using the model in \cite{Cao_2017_CVPR} pretrained on the MSCOCO dataset \cite{Lin_2014_ECCV}. 

\subsection{Prediction on 2D Datasets}

\begin{table}[!tb]
\caption{Performance of the proposed algorithm on PIE and JAAD datasets.} 
\resizebox{\columnwidth}{!}{%
\begin{tabular}{ll|cccc|cccc}
\multicolumn{2}{l}{}                & \multicolumn{4}{|c|}{PIE} & \multicolumn{4}{c}{JAAD} \\ \cline{3-10} 
\multicolumn{2}{c|}{Models}         & Acc  & AUC  & F1  & Prec & Acc  & AUC  & F1  & Prec  \\ \hline
\multicolumn{2}{l|}{TF}             & 0.75 &0.73&0.61&0.55 &  0.76&	0.72&	0.54&	0.40  \\
\multicolumn{2}{l|}{I3D \cite{Carreira_2017_CVPR}}        & 0.63 &0.58&0.42&0.37 &  0.79&	0.71&	0.49&	0.42 \\
\multicolumn{2}{l|}{MS-GRU \cite{Aliakbarian_2018_ACCV}}     & 0.86 &0.85&0.77&0.71 &  0.81&	0.79&	0.59&	0.48    \\
\multicolumn{2}{l|}{S-GRU \cite{Yue_2015_CVPR}}      & 0.82 &0.78&0.68&0.68 &  0.83&	0.76&	0.58&	0.53  \\
\multicolumn{2}{l|}{SF-GRU \cite{Rasouli_2019_BMVC}}     & 0.87 &0.85&0.78&0.74 &  0.83&	0.79&	0.59&	0.50  \\ \hline
\multicolumn{1}{c}{} & Sequential   & 0.87 &0.86&0.78&0.73 &  0.79&	0.79&	0.57&	0.45    \\
\textbf{Ours}        & Atrous        & \textbf{0.89} &\textbf{0.88}&\textbf{0.81}&\textbf{0.77} &  \textbf{0.84}&	\textbf{0.80}&	\textbf{0.62}&	\textbf{0.54}    \\
                     & Multi-scale   & 0.88&	0.86&0.79&0.75  &  
                     				 0.77&	0.79&	0.56&	0.42     
\end{tabular}
\label{results_2D}
}
\vspace{-0.4cm}
\end{table}

We begin by evaluating the performance of the proposed models and state-of-the-art algorithms discussed earlier on 2D datasets. As shown in Table \ref{results_2D}, the proposed algorithm using atrous map processing achieves the best performance on both datasets. On PIE, using multi-scale map encoding technique achieves only small improvements while the atrous method improves more on all metrics, especially on  precision by $4\%$. On JAAD, only the atrous method offers improvements on all metrics and the improvement gap is not as high as reported on PIE. This difference is to some extent expected as JAAD is not as diverse compared to PIE. For example, the majority of the samples in JAAD are pedestrians that are walking alongside the vehicle on sidewalks. This is also reflected in the changes in the performance of  I3D. This method, despite using only visual features achieves much better results on JAAD compared to PIE. This means there is a high degree of visual similarity between samples. 

Another possible explanation for smaller improvement on JAAD is the lack of proper ego-motion information. JAAD only offers high-level actions of the driver, which are not as informative as the actual dynamics of the vehicle. The importance of the ego-vehicle dynamics is highlighted in Section \ref{ablation}.

\begin{figure*}[!t]
\centering{
\includegraphics[width=0.85\textwidth]{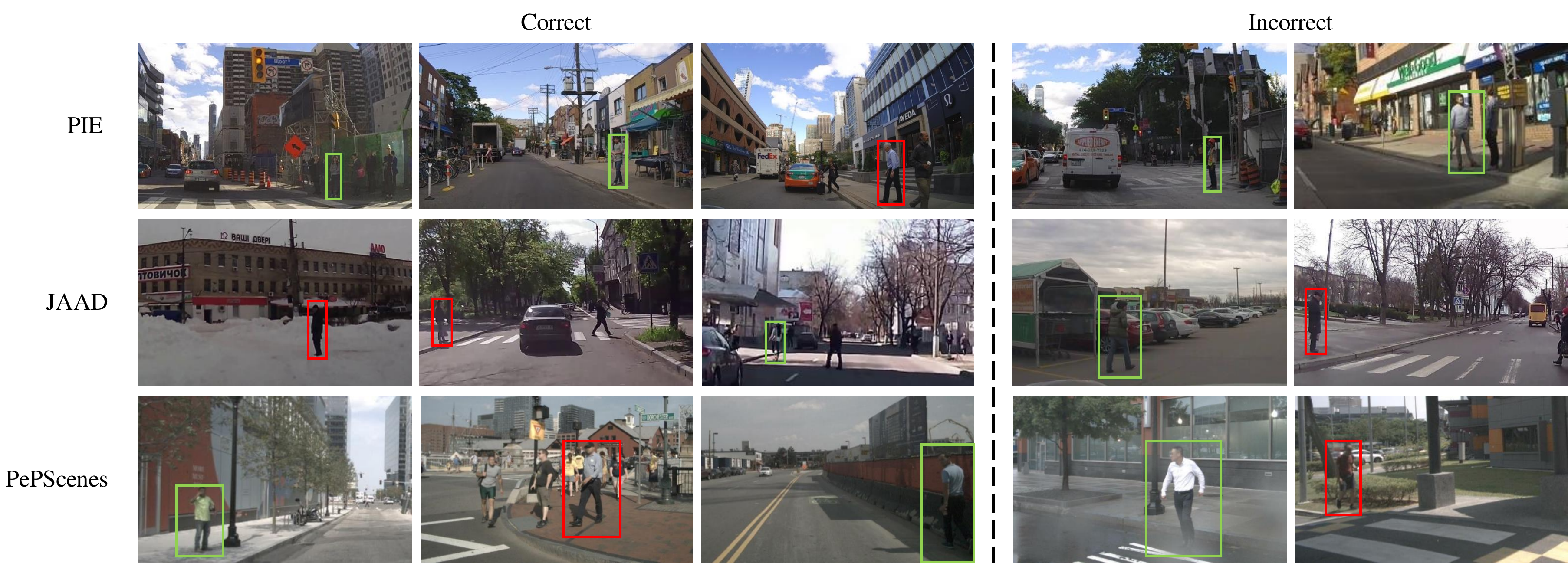}
\caption{Performance of the proposed model on 2D and 3D datasets. Red and green boxes indicate future crossing or non-crossing actions respectively. The results are divided into correct predictions on the left and incorrect ones on the right.}
\label{qualitative}}
\vspace{-0.5cm}
\end{figure*}

\subsection{Prediction on 3D Dataset}

\begin{table}[!tb]
%\resizebox{\columnwidth}{!}{%
\caption{Performance of the proposed algorithm on PePScenes.} 
\centering{
\begin{tabular}{ll|cccc}
\multicolumn{2}{c|}{Models}         & Acc & AUC & F1 & Prec \\ \hline
\multicolumn{2}{l|}{TF}           & 0.80 &0.52	&0.14	&0.26	   \\
\multicolumn{2}{l|}{I3D \cite{Carreira_2017_CVPR}}            &  0.84&	0.60&	0.33&	0.53      \\
\multicolumn{2}{l|}{MS-GRU \cite{Aliakbarian_2018_ACCV}}         &  0.85&	0.66&	0.46&	0.59    \\
\multicolumn{2}{l|}{S-GRU \cite{Yue_2015_CVPR}}          &  0.86&	0.69&	0.51&	0.64  \\
\multicolumn{2}{l|}{SF-GRU \cite{Rasouli_2019_BMVC}}         &  0.87&	0.68&	0.51&	0.67     \\ \hline
\multicolumn{1}{c}{} & Sequential   &  0.85&	0.63&	0.39&	0.65      \\
\textbf{Ours}        & Atrous       &  \textbf{0.88}&	\textbf{0.71}&	\textbf{0.56}&	\textbf{0.77}     \\
                     & Multi-scale  &  0.86&	0.68&	0.50&	0.64
    \end{tabular}}
    \vspace{-0.5cm}
\label{results_3D}
%}
\end{table}

We repeat the same experiments as above using our newly annotated 3D dataset. The results of this experiment are summarized in Table \ref{results_3D}. As in the 2D experiment, the proposed model achieves state-of-the-art performance using atrous encoding. The improvements, however, are more noticeable, in particular, F1 and precision are increased by $5\%$ and $10\%$ respectively. This is because bird's eye view maps, first, contain finer details compared to 2D semantic maps, e.g. location of objects, road boundaries, etc., and second, encode a larger area around the vehicle. As a result, avoiding downsampling and using successively larger kernel sizes can capture information more effectively.

\subsection{Qualitative Analysis}
Figure \ref{qualitative} shows the performance of the proposed model on 2D and 3D datasets. In this figure, we can see that two cases are challenging to predict: people standing on the curbside or road and have no crossing intention, e.g. the person engaged in conversation or a construction worker operating next to the road (top row). The pedestrian's direction of motion can also be distracting. For example, pedestrian might be moving towards the road or step on the road but would not cross or might be standing at the intersection prior to crossing. 

\subsection{Ablation}
\label{ablation}

Earlier we discussed the importance of multi-modal processing and how different sources of information help us capture different contextual elements in traffic scenes. Here, we conduct an ablation study on the final proposed model (with atrous encoding) using different input modalities. For this experiment, we use both a 2D dataset (PIE) and a 3D dataset (PePScenes). We selected PIE over JAAD because it has more diverse samples and it also contains actual ego-vehicle dynamics information.

Table \ref{ablation_exp} shows the results of this experiment. On the 2D dataset, the method achieves relatively high performance using only visual features. However, on all metrics, using dynamics information is superior. This is due to the fact that visual features in 2D space do not properly capture the dynamics of the scene which are fundamental for action prediction. It should be noted that despite such a discrepancy, visual features in conjunction with dynamics information can help improve the results on all metrics, particularly on F1 and precision by $4\%$ and $8\%$ respectively. 

\begin{table}[]
\caption{Ablation study on different feature modalities.} 
\label{ablation_exp}
\resizebox{\columnwidth}{!}{%
\begin{tabular}{lllll|cccc}
                                        & \multicolumn{4}{|c|}{PIE} & \multicolumn{4}{c}{PePScenes} \\ \cline{2-9} 
\multicolumn{1}{c|}{Input Modality}     & Acc  & AUC  & F1  & Prec & Acc   & AUC   & F1   & Prec   \\ \hline
\multicolumn{1}{l|}{Local Scene}              & 0.67&	0.54& 0.29&	0.37   & 0.83&	0.57&	0.27&	0.46    \\
\multicolumn{1}{l|}{Map + Local Scene}        & 0.75&	0.65& 0.49&	0.59   & 0.84&	\textbf{0.73}&	0.55&	0.53     \\
\multicolumn{1}{l|}{Ped. Motion}        & 0.84&	0.85& 0.75&	0.66   & 0.75&	0.54&	0.22&	0.23      \\
\multicolumn{1}{c|}{Ped. + Veh. Motion} & 0.85&	0.86& 0.77&	0.69   & 0.82&	0.72&	0.51&	0.47       \\ \hline
\multicolumn{1}{l|}{\textbf{Visual + Dynamics}}  & \textbf{0.89}&	\textbf{0.88}& \textbf{0.81}&	\textbf{0.77}   &
										 \textbf{0.88}&	0.71&	\textbf{0.56}&	\textbf{0.77}
\end{tabular}
}
 \vspace{-0.8cm}
\end{table}

The results on the 3D dataset show that similar to the 2D data, using only images of pedestrians and their surroundings, the performance is relatively poor. However, by adding the map information a significant improvement can be achieved, especially on AUC, F1, and precision. This is primarily due to the fact that the bird's eye view map captures the changes in the dynamics of the ego-vehicle and its surroundings much more effectively compared to 2D semantic maps. In fact, when all metrics are considered, the performance of the model is similar when using visual or dynamics features. Despite such similarity of performance, there are still cases that are not learned properly using only one type of feature. This is highlighted in the performance of the model combining all features. Besides the AUC metric, the combined model performs better on F1 and significantly better on accuracy by $4\%$ and precision by $24\%$. Considering all metrics, the combined model is clearly superior to others.  

\section{Conclusion}
In this work, we proposed a hybrid model for predicting pedestrian road-crossing action. Our model benefits from both feedforward and recurrent architectures for encoding different input modalities that capture both the changes in the visual appearance and dynamics of the traffic scenes. Using common 2D pedestrian behavior benchmark datasets and our newly annotated 3D dataset, we showed that our proposed model achieves state-of-the-art performance across all metrics. 

Furthermore, by conducting an ablation study on the proposed model, we showed how different sources of information can impact prediction accuracy. Our findings suggest that, even though dynamics information is dominant in predicting pedestrian behavior, visual features play a complementary role for prediction and can result in improved performance when combined with dynamics information.  

\bibliographystyle{IEEEtran}
\bibliography{refs}

\end{document}